\documentclass[conference]{IEEEtran}
\usepackage{graphicx} 
\usepackage{algorithm}
\usepackage{algpseudocode}
\usepackage{graphicx}
\usepackage{amsmath}
\usepackage{array}
\usepackage{boldline}
\usepackage{amssymb}
\usepackage{tikz,graphicx,xcolor}
\usepackage{gensymb}
\usepackage{textcomp}
\usepackage{float}
\usepackage{verbatim,array}
\usepackage{algorithm,algpseudocode,breqn,amsmath}
\usepackage{adjustbox,diagbox,threeparttable,tablefootnote,colortbl}
\usepackage{booktabs,tabularx,tabu,multirow,multicol,hhline}
\usepackage{caption,subcaption}
\PassOptionsToPackage{hyphens}{url} 
\usepackage{hyperref}
\usepackage[capitalise,noabbrev]{cleveref} 
\crefname{appsec}{Appendix}{Appendices} 
\usepackage[misc]{ifsym}
\usepackage[utf8]{inputenc}
\usepackage{booktabs}
\usepackage{longtable}
\usepackage{mdframed}
\usepackage{tcolorbox}
\usepackage{multicol}
\usepackage{graphicx} 
\usepackage{multirow} 

\newcommand{\name}{\texttt{VFL-SNN }}
\begin{document}

\title{Spiking Neural Networks in Vertical Federated Learning: Performance Trade-offs}

\author{
\IEEEauthorblockN{Maryam Abbasihafshejani*, Anindya Maiti‡, Murtuza Jadliwala*}
\IEEEauthorblockA{
\textit{*Department of Computer Science, University of Texas at San Antonio, TX, USA}}
\IEEEauthorblockA{\textit{‡ Department of Computer Science, University of Oklahoma, OK, USA}
}
\IEEEauthorblockA{}
}

\maketitle
\begin{abstract}
Federated machine learning enables model training across multiple clients while maintaining data privacy. Vertical Federated Learning (VFL) specifically deals with instances where the clients have different feature sets of the same samples. As federated learning models aim to improve efficiency and adaptability, innovative neural network architectures like Spiking Neural Networks (SNNs) are being leveraged to enable fast and accurate processing at the edge.
SNNs, known for their efficiency over Artificial Neural Networks (ANNs), have not been analyzed for their applicability in VFL, thus far.
In this paper, we investigate the benefits and trade-offs of using SNN models in a vertical federated learning setting. We implement two different federated learning architectures -- with model splitting and without model splitting -- that have different privacy and performance implications. We evaluate the setup using CIFAR-10 and CIFAR-100 benchmark datasets along with SNN implementations of VGG9 and ResNET classification models. 
Comparative evaluations demonstrate that the accuracy of SNN models is comparable to that of traditional ANNs for VFL applications, albeit significantly more energy efficient.
\end{abstract}

\section{Introduction}
\label{sec:introduction}
Federated learning represents a significant advancement in collaborative machine learning, allowing multiple clients to train a shared model without exchanging their local data, thus preserving data privacy~\cite{li2021survey,lu2020federated,liu2021keep,hao2019efficient}. Vertical Federated Learning (VFL)~\cite{liu2024vertical} is a special form of federated learning, particularly suited for scenarios where different clients possess different (often, non-overlapping) sets of features for the same samples. This model of learning is crucial in environments where data cannot be centralized due to privacy concerns or regulatory requirements, such as in finance, healthcare, and telecommunications. However, while VFL has been extensively studied using traditional Artificial Neural Networks (ANNs)~\cite{basheer2000artificial}, there has been limited exploration into its integration with neuromorphic and potentially more efficient models, such as Spiking Neural Networks (SNNs).

Neuromorphic or neuro-inspired learning techniques are increasingly recognized as a powerful and energy-efficient alternative to traditional deep learning methods, particularly in edge computing applications~\cite{vitale2022neuromorphic,covi2021adaptive,yoon2023environmentally,zhang2020fully}. The development and deployment of neuromorphic hardware, such as IBM's TrueNorth~\cite{akopyan2016design} and Intel's Loihi~\cite{davies2018loihi}, have propelled their popularity even further.
Among these learning techniques, SNNs have emerged as a notable candidate for delivering performance and accuracy on par with conventional deep learning algorithms while being more energy and computationally efficient~\cite{tavanaei2019deep}. Unlike classical neural networks that use continuous signal processing, SNNs operate with Integrate-and-Fire neurons that process information through discrete spikes, closely mimicking the electrical activity of neurons in the human brain.
These characteristics make SNNs an attractive option for deployment in federated learning scenarios where computational resources and energy efficiency are critical constraints. However, despite their advantages, the application of SNNs in VFL settings remains largely unexplored. This gap in research motivates our investigation into the feasibility and performance of SNNs within a VFL framework.

In this paper, we integrate SNN models into vertical federated setups and adapt existing techniques to effectively train SNNs in these settings. We design two distinct architectures for training SNN models in the vertical federated learning setting (VFL-SNN): one with model splitting, where different segments of the SNN are processed and stored across different clients, and another without model splitting, where the entire model, albeit distributed, is collaboratively updated without segmenting the model. Each architecture offers different benefits and trade-offs in terms of privacy preservation and computational efficiency and is adaptable to various operational needs and privacy requirements.

We evaluate the performance of 
the VFL-SNN models using well-known benchmark datasets, CIFAR-10 and CIFAR-100, and adapting two popular convolutional neural network (CNN) models, VGG9 and ResNET, for use with SNNs. Our comparative analysis focuses on several key metrics, including model accuracy and energy consumption, comparing these to results obtained with traditional ANNs in similar VFL setups. Our results indicate that  VFL-SNN models achieve comparable accuracy to traditional ANNs while significantly reducing energy consumption, thereby confirming the potential of SNNs in vertical federated learning environments. This research not only extends the current understanding of federated learning applications, but also demonstrates potential for utilizing energy-efficient neural network models in privacy-sensitive collaborative learning tasks.

\section{Background and Related Work}
\label{sec:background}
This section presents an overview of federated machine learning models, SNNs, and prior work in the union of these fields.
\subsection{Spiking Neural Networks (SNNs)}

A SNN consists of three main components: an \emph{encoder} that transfers input data to a temporal dimension, the \emph{SNN model}, which processes this data by leveraging the dynamics of synaptic interactions and neuronal firing patterns, and a \emph{decoder} that translates the output of  SNN from the spike data back to a form that can be understood or used in the context of the problem being solved. Following is a brief explanation of the individual components:
\subsubsection{\textbf{Input Encoding}}
Input encoding transforms information into \emph{spike trains}, allowing it to effectively represent input content as discrete spikes in a time-series. A common method for this transformation is \emph{rate coding}, which quantifies information by averaging the number of spikes over a parameterized time period. Rate coding is particularly effective in image processing tasks like classification, where it converts pixel intensity into firing rates, successfully capturing the complex details of images~\cite{kim2018deep}. 
In this encoding method, the frequency of the spikes is proportional to the magnitude of the input value. For instance, a higher pixel intensity in an image can result in a higher firing rate of spikes.
\emph{Poisson-distributed spike train}~\cite{heeger2000poisson} is a widely used type of rate encoding , where spikes are generated at each time step following a Poisson process,
in which the probability of spike occurrence is typically linked to the intensity or rate of the input signal.
A limitation of rate encoding is its disregard for the timing of spikes. To overcome this, several temporal encoding methods have been introduced. The \emph{time-to-first-spike method}~\cite{kim2018deep} is a notable technique that considers the timing of the first spike within a specific interval as the encoded information. \emph{Phase encoding} utilizes the phase of a pulse relative to a background oscillation to mark the arrival time of a spike. Additionally, \emph{correlation} and \emph{synchrony encoding}~\cite{tanoni2019spiking} are used to represent spatio-temporal spike patterns in response to specific stimuli.
\subsubsection{\textbf{SNN Models}}
Several SNN models have been developed in the research literature, with three of the most commonly used models outlined below:
\begin{itemize}
\item \textbf{Leaky Integrate-and-Fire Model:} This model is one of the simplest and most widely used in SNN architectures. In this model, neurons receiving inputs (post-synaptic neurons) aggregate the action potentials from sending neurons (pre-synaptic neurons). If the accumulated membrane potential surpasses a certain threshold, the neuron generates a spike. This spike then travels through synaptic connections, influencing the membrane potential of connected neurons~\cite{tanoni2019spiking}. The equation governing the Leaky Integrate-and-Fire (LIF) model is as follows:
\[
\tau \frac{dv(t)}{dt} = -(v(t) - v_{\text{rest}}) + R \cdot I(t)
\]
where $\tau = R \cdot C$ is the membrane time constant, indicating how quickly the membrane potential responds to changes in input current, \(v(t)\) is the membrane potential at time \(t\), \(v_{\text{rest}}\) is the resting membrane potential, \(R\) is the membrane resistance, and \(I(t)\) is the input current at time \(t\).
\item \textbf{Spike Response Model:} The Spike Response Model (SRM) incorporates the refractory dynamics of a neuron following a spike. This model reflects the temporary unresponsiveness of the neuron after firing, during which it undergoes a refractory period before it can return to its baseline voltage level. Several factors contribute to this state: the neuron might show reduced responsiveness, an elevated firing threshold, or a membrane potential that drops below the normal resting level~\cite{tan2020spiking}.
\item \textbf{Izhikevich Model: }
The Izhikevich model is designed to simulate the dynamics of a neuron's membrane potential along with its recovery characteristics~\cite{tanoni2019spiking}. Unlike other models with a fixed firing threshold, the Izhikevich model allows the threshold to vary depending on the recent activity of the neuron. This feature makes the model highly versatile and capable of representing various neuron types found throughout the brain, accurately reflecting their complex behavior.
\end{itemize}
\subsubsection{\textbf{Output Decoding}}
Decoding in SNN serves as the counterpart to encoding, converting spike trains back into interpretable information. Methods such as rate-based decoding directly translate neuron firing rates into input intensities, aligning with rate encoding techniques. 
Temporal decoding strategies reconstruct precise input timing from spikes, complementing temporal encoding like time-to-first-spike. 
Population decoding uses collective neuronal activity to infer input characteristics, mirroring population encoding techniques~\cite{dayan2005theoretical}.
\subsection{Batch Normalization Through Time (BNTT)}
The non-differentiable nature of a spiking neuron makes accurate and efficient training of SNNs challenging. To overcome this training issue in SNN, \cite{kim2021revisiting} proposes an adapted batch normalization approach that operates in the temporal dimension, called the \emph{Batch Normalization Through Time (BNTT)}. BNTT captures the temporal dynamics of spikes during batch normalization by associating a learnable parameter ($\gamma$) to each time step. These temporally evolving learnable parameters in BNTT allow a neuron to control its spike rate through different time steps, thereby expanding the batch normalization layer through time and enabling accurate and efficient training from scratch. During \emph{forward propagation}, the BNTT layer is applied after each convolutional or linear layer as follows:
\begin{align}
\label{eqn:fp}
    v_i^t & =  \lambda v_i^{t-1} + BNTT_{\gamma_i^t} (\sum_{j \in n} w_{ij} o_j^t) \nonumber \\
      & =  \lambda v_i^{t-1} + \gamma_i^t (\frac{\sum_{j \in n} w_{ij} o_j^t - \mu_i^t}{\sqrt{(\sigma_i^t)^2 + \epsilon}})
\end{align}
where, $\mu_i^t$ and $\sigma_i^t$ are the mean and variance from samples in a mini-batch $b$ at time instant $t$. $o_j^t$ is the (binary) output of neuron $j$ at time $t$ which takes a value 1 when $j$ fires at time $t$ and 0 otherwise, $v_i^t$ is the membrane potential of neuron $i$ at time $t$, $\lambda$ is the leak factor and $n$ is the set of input neurons connected to neuron $i$. The parameter $\gamma_i^t$, associated with each neuron and different for each time instant $t$, is learnt during backpropagation. During testing/validation, a global value for both mean and variance is used, which are determined by computing a exponential moving average over the values estimated during the training rounds.
\subsection{Federated Learning}

Federated machine learning offers a framework where different entities can collaboratively train a model without having to share their data. Here, we  provide an overview of two main types of federated learning, differentiated by their approach to data partitioning: \emph{Horizontal Federated Learning (HFL)} and \emph{Vertical Federated Learning (VFL)}.

\subsubsection{\textbf{Horizontal Federated Learning}}

In HFL, participants have datasets that share the same feature space but differ in the data samples. An example of this is various hospitals possessing similar medical record formats for different patients. These entities are typically unable to share their data due to privacy regulations. However, they can engage in HFL by sharing only their model updates with a server. The server orchestrates the learning process, selecting which participants contribute in each round and updating the global model using algorithms like FedAvg~\cite{sun2022decentralized}, FedProx~\cite{li2020federated}, and FedMA~\cite{wang2020federated}.

\subsubsection{\textbf{Vertical Federated Learning}}

\begin{figure}[t]
    \centering
    \includegraphics[width=0.99\linewidth]{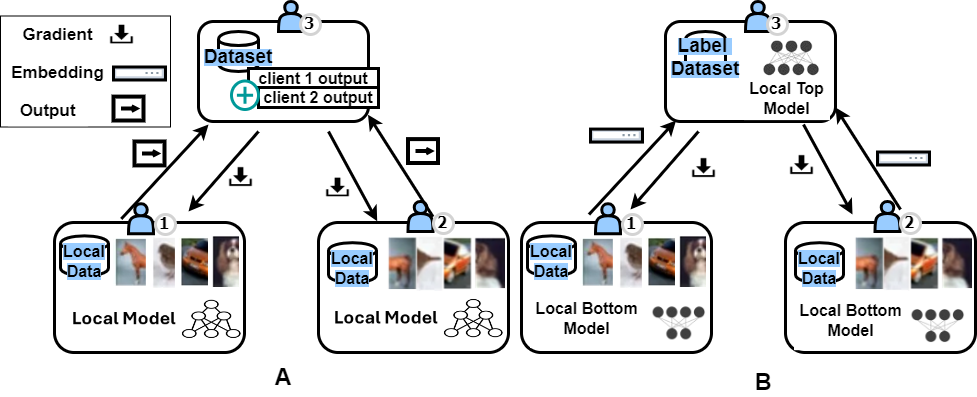}
    \caption{VFL (A) without model splitting, and (B) with model splitting.}
    \label{fig:vfloverview}
\end{figure}

VFL is applicable in scenarios where participants hold data on the same subjects but the data features vary. For example, a scenario might involve a consortium of smart home device manufacturers collaborating to improve energy efficiency and enhance user comfort using Vertical Federated Learning. Each manufacturer gathers proprietary data on energy consumption patterns, indoor climate conditions, and user preferences. Due to privacy regulations, these manufacturers cannot directly share their data. Instead, they use VFL techniques to collectively train a model that predicts optimal energy usage and adjusts home settings autonomously, ensuring privacy while optimizing smart home functionalities. VFL is categorized into two types (\cref{fig:vfloverview}): (i) \emph{without model splitting} and (ii) \emph{with model splitting}~\cite{thapa2022splitfed}. In scenarios without model splitting, each participant runs a complete model on their data and sends the results to a server. The server then aggregates these outputs to derive the final model output. Based on this aggregation, the server updates the gradients for each participant. In the with model splitting scenario, the machine learning model is segmented at a designated ``cut layer'', creating a top model and multiple bottom models (one for each participant). Each participant processes local data through their bottom model, while the top model, usually managed by a participant with access to labels, aggregates these inputs to compute the final output. The server then updates each participant’s model based on the computed gradients.

\subsection{SNN and Federated Learning}

Recent research works in the literature have explored different approaches to implement federated learning in SNNs. The application of HFL within SNNs has been implemented with adaptations made to standard SNN learning parameters to suit a distributed dataset environment~\cite{venkatesha2021federated}. There have also been efforts to enhance the training of deep SNNs, particularly emphasizing the need to refine techniques like batch normalization to boost performance and decrease network latency~\cite{kim2021revisiting}. The concept of asynchronous federated learning has been introduced to increase the efficiency and scalability of neuromorphic learning in SNNs, accommodating the delays and computation styles typical of neuromorphic systems~\cite{wang2023efficient}. Additionally, innovations in network design, such as integrating slimmable neural networks with superposition coding, are being explored to improve the data efficiency and adaptability of federated learning frameworks~\cite{yun2023slimfl}. 
\emph{Despite this progress, there is a noticeable lack of research exploring the implementation of VFL with SNNs}. To bridge this gap, we adapt the vertical federated learning training algorithm for SNN models, exploring both split and non-split model architectures. We then investigated the advantages and trade-offs of utilizing SNNs these setups and compared them with those of using ANN models.
\section{Vertical Federated Training of Spiking Neural Networks}
The first step to efficiently training SNNs is to prepare the input data by appropriately encoding it. 
As discussed in \cref{sec:background}, the choice of encoding algorithm depends on the type of data being used. In this work, as we explore application of VFL with image data, and accordingly employ a rate encoding technique. Rate encoding works well for images because it effectively captures image features by using pixel intensity as the rate of spikes.

We assume a VFL system comprising of $K$ participants/clients $\{P_k\}_{k=1}^K$ 
and a server $S$.
One of the participant can act as the server, however for ease of exposition we treat them as separate entities here. The system assumes a dataset comprising of $N$ (overlapping) samples $\mathcal{D} \triangleq \{(\mathbf{x}_i, y_i)\}$. In a VFL setting, all the participants $P_i$ and the server $S$ attempt to collaboratively train a joint machine learning model. Each feature vector $\mathbf{x}_i \in \mathbb{R}^{1,d}$ in the dataset $\mathcal{D}$ is distributed (in a non-overlapping fashion) among the $K$ participants $\{ \mathbf{x}_{i, k} \in \mathbb{R}^{1,d_k} \}_{k=1}^K$, where $\mathbf{x}_{i, k}$ and $d_k$ is the feature (partition) held by participant $P_k$ and its dimension, respectively.

As discussed earlier, a VFL system can be trained in two ways, depending on whether the top (or global) model is a trainable learning model or simply an aggregation function. The former one is referred to as VFL with model splitting, while the latter one is referred as VFL without model splitting. 
Next, we first describe VFL-SNN training approach with model splitting (outlined in \cref{alg:vflsnn-wsplit}). Following that, we describe the approach without model splitting (outlined in \cref{alg:vflsnn-wosplit}).

\subsection{VFL-SNN with Model Splitting}
In VFL with model splitting, each participant $k$ trains a local model $\mathcal{G}_k$, parameterized by $\theta_k$, using its corresponding local data (i.e., $\mathbf{x}_{i, k}$). These local participant models are also referred to as \emph{bottom models}.  
Additionally, a global model $\mathcal{F}$ (also referred to as the \emph{top model}), parameterized by $\psi$, operates on the server. The global or top model uses the outputs of the local models $\mathcal{G}_k$ to compute the final prediction by minimizing some joint task loss $\mathcal{L}$:

\begin{equation}
\begin{split}
\label{eq:withS_loss}
\small
\mathcal{L} = Loss\_Function(\mathcal{F}(\psi;   \text{Concat}(o_1(\theta_1; x_{i,1}), \ldots\\, o_K(\theta_K; x_{i,K}))), y_i)
\end{split}
\end{equation}

Where $Loss\_Function()$ is some loss function such as cross-entropy. 
$o_k(\theta_k; x_{i,k})$ represents the output of the $k$-th bottom model for the $i$-th data sample, parameterized by $\theta_k$.
$\texttt{Concat}()$ represents the concatenation of outputs from all bottom models.
Then, the server computes the gradient of this loss with respect to the top model parameters, i.e., $\frac{\partial \mathcal{L}}{\partial \psi}$, by unrolling along the time dimension. This gradient is used to update the top model parameters using the learning rate $\eta_2$, i.e., 
$\psi = \psi - \eta_2 \times \frac{\partial \mathcal{L}}{\partial \psi}$.
Simultaneously, the gradient of the loss with respect to each of the outputs from the bottom models $\frac{\partial \mathcal{L}}{\partial o_k}$ is computed. These gradients are sent back to each participant, who then performs their own backpropagation. Each participant calculates the gradient of the loss with respect to their model parameters $\frac{\partial \mathcal{L}}{\partial \theta_k}$ using this received gradient $\frac{\partial \mathcal{L}}{\partial o_k}$ (from the server).
This involves using the following chain rule  $\frac{\partial \mathcal{L}}{\partial \theta_k} = \frac{\partial \mathcal{L}}{\partial o_k} \times \frac{\partial o_k}{\partial v_k} \times \frac{\partial v_k}{\partial \theta_k}
$ and unrolling the network along the time dimension. As $o_k$ is a thresholding function, its derivative is not defined at the time of the spike and is zero everywhere else, which makes computing the gradient $\frac{\partial \mathcal{L}}{\partial \theta_k}$ intractable. This can be addressed by approximating  $\frac{\partial o_k}{\partial v_k}$ using either a piece-wise linear or exponential function \cite{neftci:2019}.
Once the gradient $\frac{\partial \mathcal{L}}{\partial \theta_k}$ is approximated, each participant updates their model parameters by applying 
$\theta_k = \theta_k - \eta_1 \times \frac{\partial \mathcal{L}}{\partial \theta_k}$,
where $\eta_1$ is the learning rate for the bottom models.

This structure allows the participants to train complex SNN models collaboratively in a VFL setting with the top layers of the model being trained centrally on the server, thereby reducing exposure to raw data. A significant security advantage in this setup is that only the server has access to the labels, enhancing privacy protection.
\begin{algorithm}
\scriptsize
\caption{Training VFL-SNN  with Model Splitting.}
\label{alg:vflsnn-wsplit}
\begin{algorithmic}[1]
\State  \textbf{Input:}
\State \hskip 1em  Dataset: $\mathcal{D}$ (participant $P_k$ privately holds $\mathbf{x}_{i, k}$, server $S$ privately holds $y_i$)
\State \hskip 1em  Hyperparameters: learning rates ($\eta_1$ and $\eta_2$ for bottom and top models, respectively), number of epochs ($E$), number of time stamps ($T$)
\State \textbf{Output: } 
\State \hskip 1em Trained local/bottom models $\mathcal{G}_k$ defined by parameters $\theta_k$ ($\forall k=1 \ldots K$) and trained global/top model $\mathcal{F}$ defined by parameter $\psi$
\item[]
\Procedure{\texttt{VFL-SNN()}}{}
\State Participants $P_1, P_2, \ldots, P_K$ initialize model parameters $\theta_1, \theta_2, \ldots, \theta_K$, respectively, and server $S$ initializes parameter $\psi$ with random weights.
	\While {stopping epoch ($E$) not met} 
		\For{each batch $b$ of samples}
			\For{each participant $P_k, k \leftarrow$ $1$ to $K$} \Comment{can be executed in parallel}
				\State $o_k \leftarrow$ perform \texttt{BNTT-FPropagateParticipant($\mathcal{G}_k$)} 
			\EndFor
			\State $o_{all}$ $\leftarrow$ \texttt{Concat}($o_1, o_2, \ldots, o_K$) \Comment{concatenate local/bottom model outputs}
			\State $o_{final} \leftarrow$ perform \texttt{BNTT-FPropagateServer($\mathcal{F}$,$o_{all}$)} 
			\State $\mathcal{L}$ $\leftarrow$ $Loss\_Function(o_{final}, y)$  \Comment{as per Equation \ref{eq:withS_loss}}
			\State $S$ computes $g_{top}^{S} \leftarrow \frac{\partial{\mathcal{L}}}{\partial{\psi}}$	\Comment{compute loss and backpropagate top model gradient}
			\State $S$ updates $\psi \leftarrow \psi - \eta_{2} g_{top}^{S}$ \Comment{update the top model}
			\For{each $k \leftarrow$ $1$ to $K$} 
				\State $S$ computes $g_{k} \leftarrow \frac{\partial{\mathcal{L}}}{\partial{o_k}}$ and forwards it to each $P_k$
				\State Each $P_k$ executes \texttt{ParticipantBP($\mathcal{G}_k$,$g_{k}$,$o_{k}$)}

			\EndFor
	\EndFor
	\EndWhile
\EndProcedure
\item[]
\Procedure{\texttt{BNTT-FPropagateParticipant($\mathcal{G}_k$)}}{}
	\For{time step $t \leftarrow$ $1$ to $T$}
	\State $o \leftarrow PoissonGenerator(b)$ \Comment{Convert pixels to spike trains}
		\For{$layer$ $\leftarrow$ $1$ to $L_{\mathcal{G}_k}-1$ of $\mathcal{G}_k$}
		\State Forward propagate as outlined in \cref{eqn:fp}
		\EndFor
	\State Accumulate membrane potential at the last layer ($L_{\mathcal{G}_k}$)
	\EndFor
	\State \Return the accumulated membrane potential $o_k$
\EndProcedure
\item[]
\Procedure{\texttt{BNTT-FPropagateServer($\mathcal{F}$,$o_{all}$)}}{}
		\For{$layer$ $\leftarrow$ $1$ to $L_{\mathcal{F}}-1$ of $\mathcal{F}$}
			\State Forward propagate as outlined in \cref{eqn:fp}
		\EndFor
	\State Accumulate membrane potential at the last layer ($L_{\mathcal{F}}$) into $o_{acc}$
	\State \Return $o_{final}$ $\leftarrow$ Averaging the firing rate 
\EndProcedure
\item[]
\Procedure{\texttt{ParticipantBP($\mathcal{G}_k$,$g_{k}$,$o_{k}$)}}{}
	\State $P_k$ computes $g_{bot}^{k} \leftarrow g_{k}.\frac{\partial{o_{k}}}{\partial{\theta_k}}$	\Comment{compute loss and backpropagate bottom model gradient}
	\State $P_k$ updates $\theta_k \leftarrow \theta_k - \eta_{1} g_{bot}^{k}$ \Comment{update the top model}
\EndProcedure
\end{algorithmic}
\end{algorithm}

\subsection{ VFL-SNN without Model Splitting}
VFL without model splitting involves each participant running a bottom model $\mathcal{G}_k$, but unlike model splitting, there is no trainable top model on the server. Instead, the server uses a simple aggregation function, such as \emph{summation}, to combine outputs $o_k = \mathcal{G}_k(\theta_k; \mathbf{x}_{i,k})$ from all participants' bottom models to compute the final output using a joint task loss function (such as cross-entropy), as shown below:
\begin{equation}
\label{eg:nonS_loss}    
\mathcal{L} = Loss\_Func\left(\sum_{k=1}^K o_k, y_i\right)
\end{equation}
This approach minimizes the computational burden on the server and reduces the potential for sensitive data exposure, as the server merely aggregates model outputs without further processing or learning. 
The server then computes the gradient of this joint loss ($\mathcal{L}$) with respect to each output ($o_k$), i.e., $\frac{\partial{\mathcal{L}}}{\partial{o_k}}$,
which are sent back to each corresponding participant in order to inform them how their local model parameters should be adjusted. 
Each participant uses these gradients ($\frac{\partial{\mathcal{L}}}{\partial{o_k}}$) received from the server to backpropagate, as described in the chain rule for the model splitting case, to calculate the gradient of the loss with respect to its own model parameters, i.e., $\frac{\partial{\mathcal{L}}}{\partial{\theta_k}}$. Participants finally update their model parameters $\theta_k$ based on these gradients to optimize their local models.
\begin{algorithm}
\scriptsize
\caption{Training VFL-SNN without Model Splitting.}
\label{alg:vflsnn-wosplit}
\begin{algorithmic}[1]
\State  \textbf{Input:}
\State \hskip 1em  Dataset: $\mathcal{D}$ (participant $P_k$ privately holds $\mathbf{x}_{i, k}$, server $S$ privately holds $y_i$)
\State \hskip 1em  Hyperparameters: learning rate ($\eta$), number of epochs ($E$), number of time stamps ($T$)

\State \textbf{Output: } 
\State \hskip 1em Trained local model $\mathcal{G}_k$ defined by parameters $\theta_k$ ($\forall k=1 \ldots K$)
\item[]
\Procedure{\texttt{VFL-SNN()}}{}
\State Participants $P_1, P_2, \ldots, P_K$ initialize model parameters $\theta_1, \theta_2, \ldots, \theta_K$, respectively.

	\While {stopping epoch ($E$) not met} 
		\For{each batch $b$ of samples}
			\For{each participant $P_k, k \leftarrow$ $1$ to $K$} \Comment{can be executed in parallel}
				\State $o_k \leftarrow$ perform \texttt{BNTT-FPropagateParticipant($\mathcal{G}_k$)} 
			\EndFor
			\State $o_{sum}$ $\leftarrow$ $\sum_k(o_k)$ \Comment{aggregate local model outputs}
			\State $S$ computes $\mathcal{L}$ $\leftarrow$ $Loss\_Function(o_{sum}, y)$  \Comment{as per Equation \ref{eg:nonS_loss}}
			\For{each $k \leftarrow$ $1$ to $K$} 
				\State $S$ computes $g_{k} \leftarrow \frac{\partial{\mathcal{L}}}{\partial{o_k}}$ and forwards it to each $P_k$
				\State Each $P_k$ executes \texttt{ParticipantBP($\mathcal{G}_k$,$g_{k}$,$o_{k}$)}

			\EndFor
	\EndFor
	\EndWhile
\EndProcedure
\item[]
\Procedure{\texttt{BNTT-FPropagateParticipant($\mathcal{G}_k$)}}{}
	\For{time step $t \leftarrow$ $1$ to $T$}
	\State $o \leftarrow PoissonGenerator(b)$ \Comment{Convert pixels to spike trains}
		\For{$layer$ $\leftarrow$ $1$ to $L_{\mathcal{G}_k}-1$ of $\mathcal{G}_k$}
		\State Forward propagate as outlined in \cref{eqn:fp}
		\EndFor
	\State Accumulate membrane potential at the last layer ($L_{\mathcal{G}_k}$)
	\EndFor
	\State \Return the accumulated membrane potential $o_k$
\EndProcedure
\item[]

\Procedure{\texttt{ParticipantBP($\mathcal{G}_k$,$g_{k}$,$o_{k}$)}}{}
	\State $P_k$ computes $g_{loc}^{k} \leftarrow g_{k}.\frac{\partial{o_{k}}}{\partial{\theta_k}}$	\Comment{backpropagate bottom model gradient}
	\State $P_k$ updates $\theta_k \leftarrow \theta_k - \eta g_{loc}^{k}$ \Comment{update the top model}
\EndProcedure
\end{algorithmic}
\end{algorithm}

\section{Evaluation}
In this section, we evaluate \name~by analyzing its energy consumption and performance under various settings.
\subsection{Experimental Setup}
We evaluated \name~using the CIFAR-10 and CIFAR-100 datasets \cite{alex2009learning}, which are commonly employed as benchmarks in image classification tasks. Both CIFAR-100 and CIFAR-10 datasets comprise 50,000 32$\times$32 RGB images for training and 10,000 images for testing. Our evaluations were conducted on an NVIDIA V100 GPU equipped with 32 GB of RAM. 
In particular, we compared the performance of ANN  
and SNN implementations using the VGG9~\cite{simonyan2014very} and ResNet-18~\cite{koonce2021resnet} architectures in vertical federated learning setups. 
The VGG9 model is designed to effectively process and classify complex visual data, utilizing seven convolutional layers with 3$\times$3 filters and two linear layers. The ResNet-18 model is a robust convolutional neural network that integrates 18 layers, featuring convolutional layers and skip connections that allow bypassing of certain layers. This design enables the network to learn from both shallow and deep layer features, thereby enhancing the training efficiency of the network.
We distributed the features vertically among $N$ clients and conducted experiments using \name~with both model splitting and without model splitting. \name~implementation was derived from the HFL-SNN code base of ~\cite{venkatesha2021federated}, with significant additions and modifications made for the VFL setup. 
We used  the Leaky Integrate-and-Fire (LIF) neuron model with a neuron leakage rate ($\lambda$) set at 99\% and Poisson encoding~\cite{heeger2000poisson} for encoding inputs.
Poisson encoding is a type of rate encoding that introduces unpredictability and temporal variations in spike patterns, enhancing the network's adaptability to diverse scenarios and its responsiveness to noisy inputs. To decode the output from SNN models, we compute the average firing rate of the output neurons. For classification tasks, these firing rates are averaged, and the neuron exhibiting the highest firing rate is identified as representing the predicted class. For the \name~VGG9 model, we chose a spike learning rate of $0.1$, a leak memory rate of $0.99$, a momentum of $0.95$, and a weight decay of $1 \times 10^{-4}$, with the Stochastic Gradient Descent (SGD) optimizer employed to optimize the model. In the \name~ResNet18, the best accuracy is obtained with a learning rate of $0.02$, while the other parameters are left unchanged. Due to the differences in architecture of ANN models, we use a different set of optimized parameters, for comparisons with \name. For reference implementations of VFL-ANN (artificial neural network for vertical federated learning) models of VGG9 and ResNet18, we obtained the best accuracy with a learning rate of $0.001$ and a weight decay of $5 \times 10^{-4}$. The models were configured to classify 10 different classes for CIFAR-10 and 100 for CIFAR-100.
\begin{table*}[h]
\caption{Performance metrics comparison for CIFAR-10 and CIFAR-100 for VFL without model splitting.}
\label{tab:ws}
\centering
\begin{tabular}{|l|p{0.5 cm}|p{0.8 cm}|p{0.5 cm}|p{1 cm}|p{0.5cm}|p{1cm}|p{1cm}|p{1cm}|}
\hline
\textbf{Dataset} & \multicolumn{4}{c|}{\textbf{CIFAR-10}} & \multicolumn{4}{c|}{\textbf{CIFAR-100}} \\
\hline
\fontsize{6pt}{7pt}\selectfont\textbf{Metric} & \fontsize{6pt}{7pt}\selectfont\textbf{SNN-VGG9} & \fontsize{6pt}{7pt}\selectfont\textbf{SNN-ResNET18} & \fontsize{6pt}{7pt}\selectfont\textbf{ANN-VGG9} & \fontsize{6pt}{7pt}\selectfont\textbf{ANN-ResNET18} & \fontsize{6pt}{7pt}\selectfont\textbf{SNN-VGG9} & \fontsize{6pt}{7pt}\selectfont\textbf{SNN-ResNET} & \fontsize{6pt}{7pt}\selectfont\textbf{ANN-VGG9} & \fontsize{6pt}{7pt}\selectfont\textbf{ANN-ResNET18}\\
\hline
\textbf{Average Training Accuracy} & 80.38 & 88.61 & 92.79 & 75.48 & 65.35 & 65.40 & 67.32 & 66.56 \\
\textbf{Best Test Accuracy} & 77.95 & 74.10 & 80.35 & 77.09 & 50.93 & 44.01 & 49.07 & 44.50 \\
\textbf{Final Test Accuracy} & 77.64 & 72.66 & 79.6 & 75.51 & 50.41 & 42.65 & 48.72 & 43.81 \\
\textbf{Average Test Accuracy} & 71.02 & 71.23 & 77.14 & 67.65 & 44.92 & 40.13 & 44.10 & 38.85 \\
\textbf{Precision} & 0.777 & 0.740 & 0.804 & 0.774 & 0.521 & 0.449 & 0.507 & 0.4706 \\
\textbf{Recall} & 0.779 & 0.741 & 0.803 & 0.770 & 0.509 & 0.440 & 0.490 & 0.444 \\
\textbf{F1 Score} & 0.778 & 0.740 & 0.803 & 0.772 & 0.515 & 0.444 & 0.498 & 0.457 \\
\hline
\end{tabular}
\end{table*}

\begin{table*}[h]
\centering
\caption{Performance metrics comparison for CIFAR-10 and CIFAR-100 for VFL with model splitting.}
\label{tab:withs}
\begin{tabular}{|l|p{0.5 cm}|p{0.8 cm}|p{0.5 cm}|p{1 cm}|p{0.5cm}|p{1cm}|p{1cm}|p{1cm}|}
\hline
\textbf{Dataset} & \multicolumn{4}{c|}{\textbf{CIFAR-10}} & \multicolumn{4}{c|}{\textbf{CIFAR-100}} \\
\hline
\fontsize{6pt}{7pt}\selectfont\textbf{Metric} & \fontsize{6pt}{7pt}\selectfont\textbf{SNN-VGG9} & \fontsize{6pt}{7pt}\selectfont\textbf{SNN-ResNET18} & \fontsize{6pt}{7pt}\selectfont\textbf{ANN-VGG9} & \fontsize{6pt}{7pt}\selectfont\textbf{ANN-ResNET18} & \fontsize{6pt}{7pt}\selectfont\textbf{SNN-VGG9} & \fontsize{6pt}{7pt}\selectfont\textbf{SNN-ResNET} & \fontsize{6pt}{7pt}\selectfont\textbf{ANN-VGG9} & \fontsize{6pt}{7pt}\selectfont\textbf{ANN-ResNET18}\\
\hline
\textbf{Average Training Accuracy} & 83.26 &89.35 & 89.292 &83.45 & 54.26 &55.70 & 76.11&72.22 \\
\textbf{Best Test Accuracy} &80.28&77.05 &82.091 &79.47 &52.11 &48.03 & 55.61&48.80 \\
\textbf{Final Test Accuracy} & 78.89 &75.74 & 82.09 &78.01 & 51.50 &46.32 & 55.438&45.19 \\
\textbf{Average Test Accuracy} & 76.28 & 74.58 & 78.13 &74.90 & 45.02 & 42.70 & 49.28&43.64 \\
\textbf{Precision} & 0.806 &0.771 & 0.820 & 0.797 & 0.537 & 0.480 & 0.568&0.506 \\
\textbf{Recall} & 0.802 & 0.770 &0.820&0.794 & 0.521 & 0.480 & 0.556 &0.487 \\
\textbf{F1 Score} & 0.804 & 0.771 & 0.820 &0.796 & 0.528 &0.480 & 0.562&0.497 \\
\hline
\end{tabular}
\end{table*}
\begin{figure*}[]
\centering
\subfloat[CIFAR-10]{
  \includegraphics[width=0.35\linewidth]{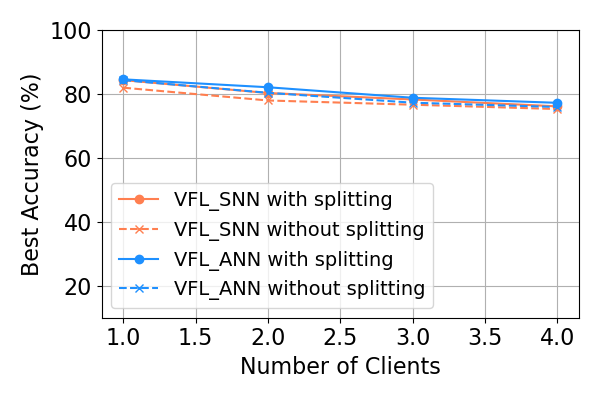}
  \label{fig:Clinet_CIFAR10}
}
\subfloat[CIFAR-100]{
  \includegraphics[width=0.35\linewidth]{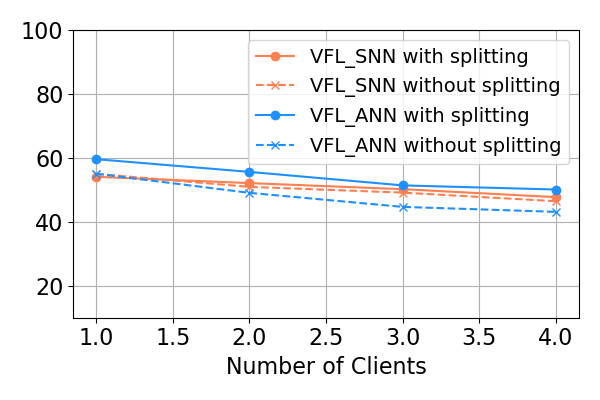}
  \label{fig:Clinet_CIFAR100}
}
\caption{Comparison of CIFAR-10 and CIFAR-100 classification accuracy across different numbers of clients.}
\label{fig:client_accuracy_comparison}
\end{figure*}

\subsection{Models and Splitting}
Our first evaluation aims to determine the impact of various learning models on performance metrics in both \name~without model splitting and with model splitting configurations. We specifically assessed the performance of both \name~and VFL-ANN versions of VGG9 and ResNet18 under conditions where there were two clients and the SNN time step was set at 32. The results, presented in \cref{tab:ws,tab:withs}, demonstrate a relatively minor performance disparity between the SNNs and ANNs versions, with about a 3\% difference. For instance, in VFL without model splitting on CIFAR-10, the best test accuracy for the ANN VGG9 is 80.35, while for the SNN VGG9, it is 77.95.
Although VFL-ANNs surpass \name~in performance, the minimal difference in accuracy can be inconsequential, positioning SNNs as a viable alternative for edge computing applications if power efficiency (evaluated in \cref{subsec:energy}) and the capability to handle time-varying data are essential.

\subsection{Number of Clients}
Next, we evaluate the models' stability by altering the number of clients and distributing segments of the image between them. For example, in a scenario with 2 clients, one client processes the right side of the images, while the other handles the left side. \cref{fig:Clinet_CIFAR10,fig:Clinet_CIFAR100} demonstrate how accuracy shifts with varying numbers of clients using the \name~VGG9 model at a fixed time-step of 32, in comparison to the ANN model across both VFL with model splitting and VFL without model splitting configurations. The results indicate that as the number of clients increases, the accuracy of the ANN model declines more sharply, suggesting that SNNs are more scalable in a vertical federated learning context. This effect is particularly pronounced with CIFAR-100 (\cref{fig:Clinet_CIFAR100}), where the accuracy of the VFL-ANN decreases from approximately 50\% with two clients to 43\% with four clients. In contrast, the accuracy for \name~drops from about 50\% to 45\% only.
\subsection{Time Steps}
\begin{figure}[t]
\centering
\includegraphics[width=0.99\linewidth]{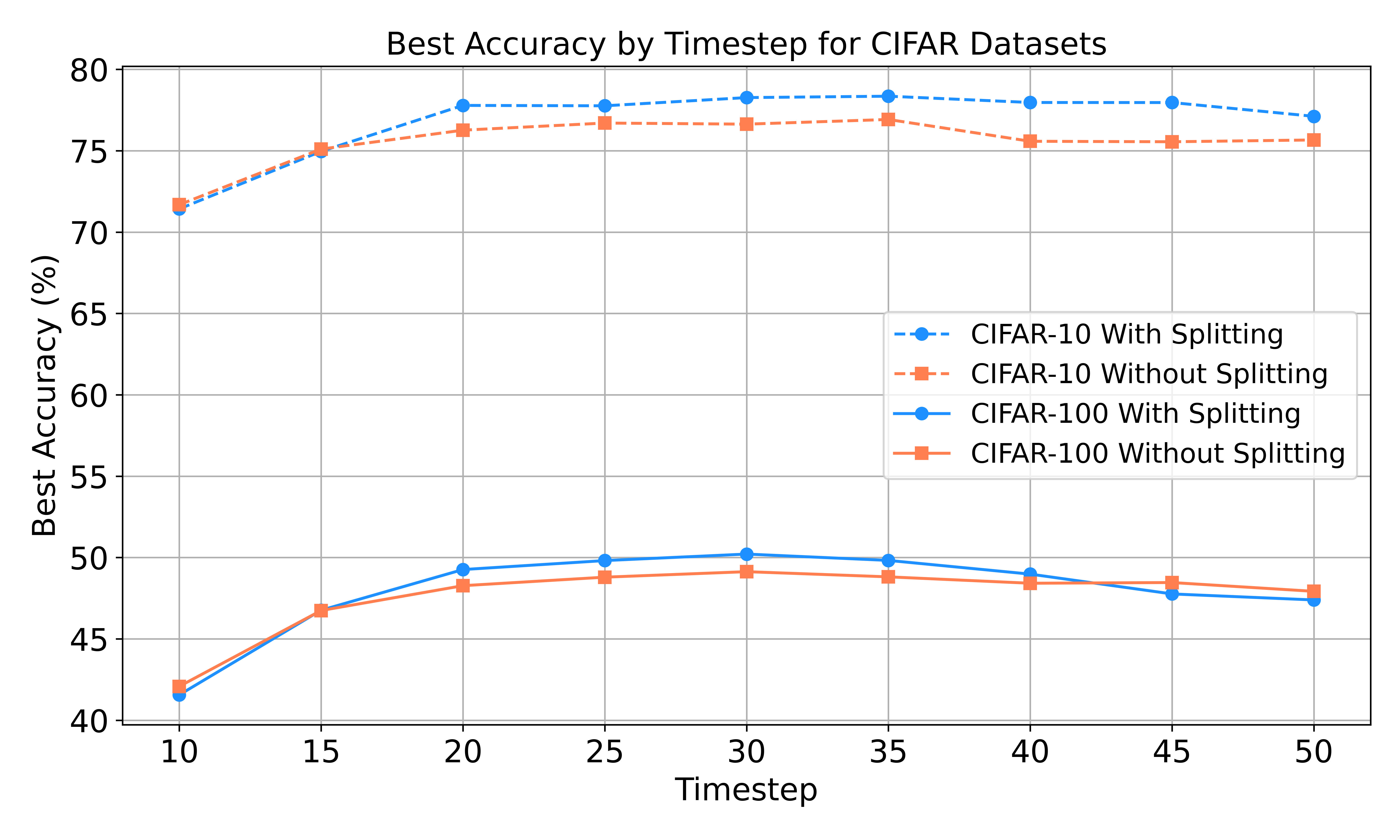}
\caption{Variations in accuracy with increasing time steps.}
\label{fig:time_CIFAR10}
\end{figure}

To examine the impact of the SNN time step on \name accuracy, we varied the time steps from 10 to 50 while maintaining three clients. As shown in \cref{fig:time_CIFAR10}, increasing the time step from 10 to about 30 led to improved accuracy. However, extending the time step beyond this point did not significantly affect performance, as evidenced by the minimal changes in accuracy.
\begin{figure}[b]
\centering
\includegraphics[width=0.99\linewidth]{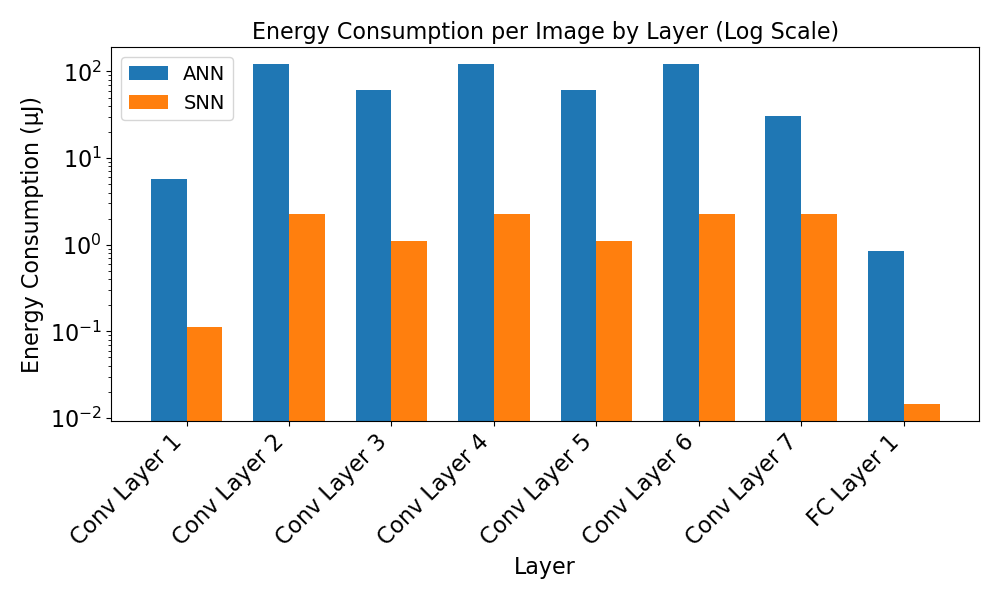}
\caption{Estimated training energy across layers for SNN (\name) and ANN VGG9 model on CIFAR-10 dataset.
}
\label{fig:energy}
\end{figure}

\subsection{Energy Consumption}
\label{subsec:energy}

In this section, we evaluate the energy consumption associated with each layer of the models during training. Because SNNs only activate a subset of neurons at any given time, energy consumption is significantly reduced. To compare the energy consumption with ANN models, we recorded the number of spikes generated by neurons in each layer of the SNN-VFL and then computed the energy consumption based on the energy required for SNN model operations as follows:
The energy required for operations (OPS) in a convolution layer  is calculated as $OPS = M^2 \times I \times k^2 \times O$, where $k$ is the kernel size, $I$ is the input channels count, $O$ is the output channels count, and $M \times M$ is the feature map size. The relationship between $M$ and the size of the input feature map, $N_I \times N_I$, leads to a refined OPS formula:
\begin{equation}
\label{eq:pos}
OPS = \left( \frac{N_I - k + 2p}{s} + 1 \right)^2 \times I \times k^2 \times O,
\end{equation}
where $p$ represents padding and $s$ is the stride. For fully connected layers, where $I$ inputs connect to $O$ outputs, the OPS formula simplifies to:
\begin{equation}
OPS = I \times O
\end{equation}
Equation~\ref{eq:pos} illustrates how the input size significantly affects energy consumption. By implementing vertical federated learning, which distributes features across multiple devices, the energy demand on each device is reduced. Additionally, SNNs utilize binary spikes that convert the multiply
accumulate (MAC) operation predominantly into an accumulation (AC) operation, greatly enhancing energy efficiency. The energy consumption for ANNs is given by:
\begin{equation}
\label{eq:ANN_energy_estimation}
E_{ANN} = OPS \times E_{MAC}
\end{equation}
Energy consumption for SNNs is based on the energy needed for AC operations, the spike rate, and time steps, and is computed as:
\begin{equation}
\label{eq:snn_energy_estimation}
E_{SNN} = OPS \times R \times T \times E_{AC}
\end{equation}
where $R$ is the average rate of spikes per layer for the entire dataset, defined by:
\begin{equation}
R = \frac{N_S}{N_d \times N_n}
\end{equation}
with $N_S$ being the number of spikes, $N_d$ the dataset size, and $N_n$ the number of neurons.
\begin{table}[h]
\centering
\caption{Energy estimations for multiply and accumulate operations \cite{venkatesha2021federated}.}
\label{tab:energy_estimation}
\begin{tabular}{|l|c|}
\hline
\textbf{Operation} & \textbf{Estimated Energy (pJ)} \\
\hline
32-bit Multiply (EMult) & 3.1 \\
\hline
32-bit Add (EAdd) & 0.1 \\
\hline
\parbox[t]{5cm}{32-bit Multiply and Accumulate \\ (EMAC = EMult + EAdd)} & 3.2 \\
\hline
32-bit Accumulate (EAC) & 0.1 \\
\hline
\end{tabular}
\end{table}
We compare the energy consumption of each layer in VFL-SNN and VFL-ANN models using equations~\ref{eq:ANN_energy_estimation} and \ref{eq:snn_energy_estimation}. To obtain the spike rate, we recorded the number of spikes during model training. To estimate the energy required for AC and MAC operations, we used the energy consumption data reported in \cite{horowitz20141} for 32-bit integer arithmetic operations in a 45nm CMOS processor, as detailed in \cref{tab:energy_estimation}.
In the estimation, the focus was on multiply and accumulate operations, excluding the energy costs associated with memory and peripheral circuits.
\cref{fig:energy} illustrates the energy consumption across different layers of the VFL-SNN and VFL-ANN VGG9 models during training. These models were trained on the CIFAR-10 dataset by three clients, with the \name's SNN model utilizing 32 time steps. The average energy consumption for ANN across all layers is approximately 64.98 microjoules ($uJ$), whereas for SNN, it is only about 1.41 $uJ$, with layers using between 0.015$uJ$ and 2.269 $uJ$. This reveals that, on average, the ANN consumes around 34.7 times more energy than the SNN. This comparison underscores the energy savings achieved through the Vertical federated learning with SNN models  demonstrating their benefit for edge devices. 

\subsection{Training Time}
\begin{figure}[h]
\centering
\includegraphics[width=0.99\linewidth]{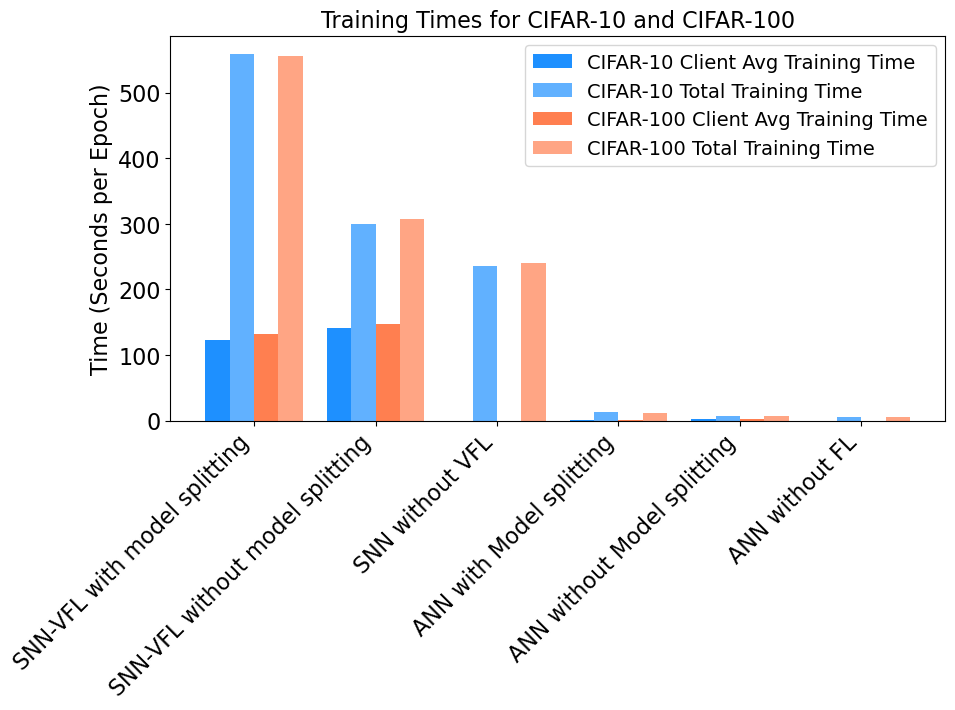}
\caption{Training time comparison across models.}
\label{fig:cifar_10_training_times}
\end{figure}

Although SNNs benefit from low energy consumption, they typically require longer training times compared to ANNs. This prolonged training period is due to the inherent use of time steps in SNNs for processing spikes. \cref{fig:cifar_10_training_times} depicts the training times with three clients and a time step of 32 for SNN VGG9 (SNN-VFL), ANN VGG9 (ANN-VFL), and also SNN and ANN without federated learning for comparison. 
Training \name~with model splitting resulted in an average client training time per epoch of 123 seconds and a total training time of 559 seconds. Without model splitting, \name had a higher average training time of 142 seconds but a much lower total time per epoch of 300 seconds. In contrast, VFL-ANN displayed dramatically lower training times of approximately 1.53 seconds per client per epoch.
It is evident that the overall training duration for \name, both with and without model splitting, is substantially longer than that for VGG9 ANNs and non-federated SNNs, underscoring the trade-off between energy efficiency and training time.

\section{Discussion and Conclusion}
\label{sec:discussion}

The integration of SNNs into VFL frameworks represents a promising direction in machine learning, particularly in scenarios demanding high computational efficiency and privacy preservation. The experimental results from the \name~framework highlight both the potential benefits and the limitations inherent in this novel approach.

\subsection{Efficiency and Performance}
The findings from our evaluation indicate that SNNs, while slightly underperforming in comparison to traditional ANNs in some metrics, offer significant improvements in energy efficiency. This trade-off is crucial in edge computing and IoT scenarios where power consumption is a limiting factor. The lower energy requirements of SNNs, as demonstrated in the \name setting, suggest that the slight decrease in accuracy is compensated by the reduced operational costs and enhanced sustainability of the models.

\subsection{Scalability and Adaptability}
The adaptability of SNN models to different VFL configurations -- namely, with and without model splitting -- offers flexible approaches tailored to varying privacy and computational constraints. The scalability of SNNs is particularly notable, as evidenced by their performance stability across different client numbers and learning configurations. This suggests that SNNs could handle increasing data or feature distribution complexities better than ANNs, a valuable characteristic as federated learning applications grow in scale and complexity. 

\subsection{Challenges and Future Work}
Despite these promising results, the extended training times of SNNs present a significant challenge. The inherent complexity of temporal data processing in SNNs, necessary for their energy efficiency and neuromorphic advantages, leads to longer training periods compared to ANNs. This aspect may limit the immediate widespread adoption of SNNs in time-sensitive applications.
Future research can focus on optimizing the training processes of SNNs within federated frameworks, perhaps by developing new algorithms that can reduce training time without compromising the model's performance or energy efficiency. Moreover, exploring more advanced SNN models and their specific adaptations for VFL could further enhance the performance metrics, closing the gap with traditional ANNs. Also, while we estimate SNN processor energy consumption based on precise approximations from spiking rate simulations and benchmarks, real-world energy use may vary due to deployment conditions, hardware differences, and variations in SNN algorithms.

\section{Conclusion}
In this study, we adjusted two VFL architectures - with model splitting and without model splitting - for Spiking Neural Networks (SNNs), addressing the challenge of efficiently conducting model training across multiple clients while preserving data privacy. We compared these adjusted \name setups with ANN-VFL setups through comprehensive evaluation on the CIFAR-10 and CIFAR-100 datasets using SNN adaptations of VGG9 and ResNet models. Our results demonstrated comparable accuracy to traditional ANNs in VFL scenarios. \name was also found to be significantly more energy-efficient, suggesting its potential for wider adoption in privacy-sensitive, distributed computing environments.

\section{Acknowledgement}
This work was supported in part by the NSF under award 2332744. The views and findings solely reflect those of the authors and should not be seen as representative opinions of any funding agency.
\bibliographystyle{plain}
\bibliography{references}

\end{document}